%% file: iclr2022_conference.tex
\documentclass{article} % For LaTeX2e
\usepackage{iclr2022_conference,times}

\input{math_commands.tex}

\usepackage{graphicx}
\usepackage{hyperref}
\usepackage{url}
\usepackage{algorithm}
\usepackage{algorithmic}
\usepackage{amsfonts}
\usepackage{dsfont}
\usepackage{balance}  % for  \balance command ON LAST PAGE  (only there!)
\newcommand{\ens}[1]{\left\{#1 \right\}}
\usepackage{newfloat}
\usepackage{listings}
\floatstyle{ruled}
\newfloat{listing}{tb}{lst}{}
\floatname{listing}{Listing}

\title{Hyperspherical embedding for novel class classification}

\author{Rafael S. Pereira \\
DEXL\\
National Laboratory of Scientific Computing\\
Petropolis  Brazil \\
\texttt{\{rpereira\}@lncc.br} \\
\And
Alexis Joly \\
Zenith INRIA \\
Mountpellier France \\
 \\
\texttt{\{alexis.joly\}@inria.fr} \\
\And
 Patrick Valduriez \\
Zenith INRIA \\
Mountpellier France \\
\texttt{\{patrick\_valduriez\}@inria.fr} \\
\And
Fabio Andre Machado Porto \\
DEXL \\
National Laboratory of Scientific Computing\\
Petropolis Brazil \\
 \\
\texttt{\{fporto\}@lncc.br} \\
}

\iclrfinalcopy % Uncomment for camera-ready version, but NOT for submission.
\begin{document}

\maketitle

\begin{abstract}
    Deep neural networks proved to be useful to learn representations and perform classification on many different modalities of data. Traditional approaches work well on the closed set problem. For learning tasks involving novel classes, known as the open set problem, the metric learning approach has been proposed. However, while promising, common metric learning approaches require pairwise learning, which significantly increases training cost while adding additional challenges. In this paper we present a method in which the similarity of samples projected onto a feature space  is enforced by a metric learning approach without requiring
pairwise evaluation. We compare our approach against known methods in different datasets, achieving results up to $81\%$ more accurate.
\end{abstract}

\section{Introduction}

Humans have the ability to identify many different types of objects, \cite{HumanVision}. Even when we are not able to name a certain object, we can tell it's differences from a second object, which contributes to the identification of objects we have never seen before and group them into  classes based on prior knowledge. \textit{Metric learning} \cite{MetricSurvey} is a well adopted approach that identifies novel classes without fine tuning a model on these classes. The approach  applies an optimization strategy, which guarantees that the classes a model has seen during optimization form disjoint clusters on the latent space according to a certain metric distance. Some common approaches that use this strategy are: the triplet loss \cite{TripletLossPaper}; constrative loss \cite{ConstrativeLossPaper}; prototypical networks \cite{PrototipicalNetworks}; constellation loss \cite{constelationloss}; and matching networks \cite{MatchingNetworkPaper} here refered to distance based learners. Another approach in metric learning is called \textit{similarity learning}, where the model receives pairs of inputs and learns that they are similar if they belong to the same class or dissimilar otherwise, as discussed in \cite{RelationNetwork}. During inference on novel classes, distance based learners uses the distance between labeled points of the novel class the model was not optimized upon to obtain a representation in the latent space for the novel class and then calculate the distance between new points and each class representation. 
When considering similarity based learners, a similarity score is calculated between every (class,query) point pair in order to find the most similar pair. 

However while enforcing metric properties on the latent space leverages the model knowledge to novel classes, it requires pairwise learning, which limits the scalability of such approaches given the amount of possible pairs.

In this paper we take into account the normalized softmax loss function(NSL), proposed by \cite{NSL}, and present how it enforces a latent space that obeys the cosine similarity. Based on this, we then present a methodology to apply the \textit{NSL} to the novel classes classification problem. Considering a trained artificial neural network, we add a new neuron to it's last layer and infer the weights that connect the penultimate layer of the network to this neuron. The connection and the new neuron are used to classify a novel class by using few labeled samples of it. Our approach for the open set problem allows us to classify new classes without fine-tuning the model, instead we use the same network parameters the model was optimized upon to classify it's seen classes and only adding a new neuron along with it's inferred connection.
We evaluate state-of-the-art approaches to solve the open set problem against our proposed approach, both in the disjoint and joint scenarios, %as defined in section \ref{sec:DisjointJointDefinition},
for different datasets. The experimental results show that our approach outperforms other metric learning strategies and additionally, induces a more scalable training process, as it does not require pairwise learning, leveraging the open set problem technique to deal with large datasets. 

The remainder of this paper is structured as follows. First, it presents some theoretical background at section Preliminaries. Our methodology and how to classify new classes is described in section Proposed Methodology. Next, we present the results on the joint and disjoint open set problem in section Results. Moreover, we present the use of the NSL approach in a more complex dataset in the field of botany, in section Case Study: The Pl@ntnet dataset. We compare our methods to incremental learning in section Few shot scenario for incremental learning. We present related work and lastly, we conclude in section Conclusion.

%\section{Problem Formalization}
\section{Preliminaries} \label{sec:Background}

We are given a training dataset $(x_i,y_i)_{i \in \ens{1,\dots,n}}$ where, for all $i$, the input $x_i$ belongs to an input space $\mathcal{X} \subset \mathbb{R}^d$, e.g. the space of images, and the output $y_i$ to an output space $\mathcal{Y} = \ens{1,2,\dots,K}$, the set of class labels, where $K$ is the number of classes.
%The joint space $\X \times \Y$ is a probabilistic space and the data points are sampled from the joint probability measure $\Pr_{X,Y}$.
%This joint probability can, in turn, be decomposed into the marginal probability measure on $\X$, denoted $\Pr_X$, and the conditional probability of $Y$ given $X$, denoted $\eta(x)$:
%\begin{equation*}
%  \eta_k(x) \eqdef \prcond{ Y=k }{ X=x }.
%\end{equation*}
%To enforce the metric property in this paper we discuss how to enforce cosine similarity in this metric space
%\hline
Based on this training set, the aim is to find a classifier $h: \mathcal{X} \to \mathcal{Y}$ which produces a single prediction for each input and generalizes well on unseen samples $x \in \mathcal{X}$. When this classifier is a deep neural network, $h$ can typically be expressed as:
\begin{equation*}
h(x)=\max_{k} \hat{\eta}_k(x)
\end{equation*}
where $\hat{\eta}(x)=(\hat{\eta}_1(x),\dots,\hat{\eta}_K(x))$ is the vector of the estimated class probabilities computed as:
\begin{equation*}
\hat{\eta}(x)=\psi(\phi(x))
\end{equation*}
with $\phi: \mathcal{X} \to \mathbb{R}^M$ being a succession of layers allowing to compute an $M$-dimensional feature vector representation $\phi(x)$ for any input image $x \in \mathcal{X}$, and $\psi: \mathbb{R}^M \to \mathbb{R}^K$ being the final classification function, typically composed of a fully connected layer followed by a softmax activation function:
\begin{equation}
\psi_k(z)=\frac{e^{w_k z + b_k}}{\sum_{j=1}^K e^{w_{j} z + b_{j}}}
\label{eq:Softmax}
\end{equation}

\subsection{The Open set problem}
The classification problem can be formulated as a closed set or open set problem. In the closed set problem context, the optimization process trains a model to learn features that can classify the samples into classes present in the training set. The approach does not require the identification of classes not present in the training set. This is commonly tackled using the Softmax-cross-entropy loss \cite{resnet},\cite{VGG},\cite{Inception}. In contrast, in the open set problem we are interested in not only identifying the classes present in the training set, but also to be able to use the model to classify new classes by exploiting properties in the latent space yielded during optimization.

\subsection{Classifying new classes}

\label{sec:NewClasses}
When tackling the open set problem, we are interested in optimizing models in which the full knowledge the network obtains during optimization can be exploited for classes outside of the training set. The usual \textit{softmax cross-entropy} approach lacks the ability to extract features that obey this property, as the weights between the penultimate layer and the classification layer $w$ are as important as the representation in the latent space of the penultimate layer $z$ as seen in equation \ref{eq:Softmax}, and the former is undefined for novel classes. Usual approaches for classifying novel classes are explored in metric learning as already discussed in the previous section. Metric learning strategies are interesting as novel classes can be defined however the presented strategies can be costly to optimize given pairwise learning. We discuss further on this paper how can we remove pairwise learning and still be able to define novel classes for an model.%  \cite{ConstrativeLossPaper},\cite{TripletLossPaper},\cite{RelationNetwork},\cite{MatchingNetworkPaper} where model optimization occurs by learning a metric space that creates clusters based on a metric distance
%\cite{ConstrativeLossPaper},\cite{TripletLossPaper},\cite{MatchingNetworkPaper}, or by learning whether two examples are similar or dissimilar \cite{RelationNetwork}. These strategies allow classification to be performed for classes outside of the training set as the representation in the latent space can be exploited via a K nearest neighbours approach. However, all these strategies are based on pairs or triplets, which limits the approaches scalability whenever optimizing with larger amounts of data per class, or a large amount of classes in a small data scenario.

\subsection{Normalized Softmax Loss}
\label{sec:NSLProp}
\begin{figure}[!h]
    \centering
    \includegraphics[width=0.4\textwidth]{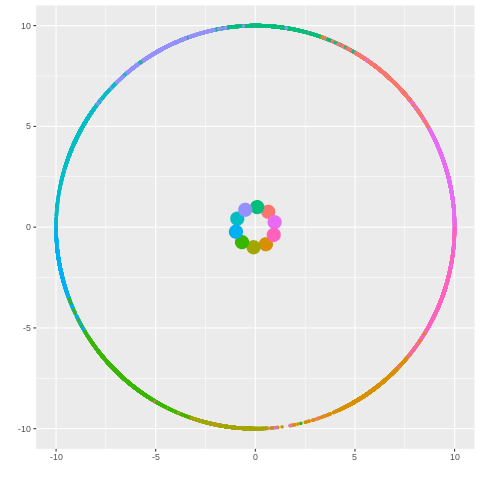}
    \caption{Embedding obtained on the cifar10 dataset when using a latent space with two dimensions using NSL, each color represents a different class. Inner points are the class weights while outer points come from the training set, notice how classes from the outer circle are aligned with the inner circle}
    \label{fig:CifarEmb2D}
\end{figure}

Proposed in \cite{NSL}, the NSL (Normalized Softmax Loss) is a modification of the \textit{softmax} loss that enforces a cosine similarity metric between classes on the latent space. It enforces the features $z$ that are projected into the latent space to be contained in a $M$ dimensional hypersphere ($M>3$) where each region of the sphere contains features belonging to a certain class.
If we look again at the classical softmax equation (Eq. \ref{eq:Softmax}), the constraints induced by NSL are:\\
\begin{equation}
\label{equ:constraints}
\left\{\begin{matrix}
b_k=0, \forall k\\ 
{\|w_{k}\|}=1, \forall k\\ 
{\|z\|}={\|\phi(x)\|}=S,\forall x\\ 
\end{matrix}\right.
\end{equation}\\
and finally
\begin{equation*}
\hat{\eta}_k(x)=\psi_k(\phi(x))=\frac{e^{w_k \phi(x)}}{\sum_{j=1}^K e^{w_j \phi(x)}}=\frac{e^{S.cos(w_k, \phi(x))}}{\sum_{j=1}^K e^{S.cos(w_j,\phi(x))}}\\ 
\end{equation*}\\
where $cos(u,v)=u.v/(\|u\|.\|v\|)$ is the cosine similarity, i.e. the cosinus of the angle between two vectors $u$ and $v$. Note that the hyper-parameter $S$ acts as a temperature of the normalized softmax allowing to control the degree of concentration of the output probabilities $\hat{\eta}_k(x)$.\\
A geometrical representation that shows the relationship between the weights and the feature vectors obtained with NSL is shown in Figure \ref{fig:CifarEmb2D}. %and \ref{fig:PlantNetEmb2D}. 
One can see that the barycenter of the feature vector is aligned with it's corresponding class weights.

\section{Proposed Methodology}
\label{sec:methodology}
In this paper we aim to compare pairwise strategies, commonly used in metric learning, against the normalized \textit{softmax} loss approach for the open set problem.
In this manner we consider both the problem where during inference seen and unseen classes are disjoint, as well as the scenario where the model must identify both the seen and unseen classes together.\\
More formally, once the network has been trained, we would like to extend the output space to a new set of classes $\mathcal{Y^*} = \ens{K+1,\dots,K+K^*}$ for which we have only one or very few samples $(x^*_i,y^*_i)_{i \in \ens{1,\dots,n*}}$. In particular, we would like to obtain a new classifier $h^*: \mathcal{X} \to \mathcal{Y^*}$ (disjoint scenario) or a new classifier $h': \mathcal{X} \to \mathcal{Y} \bigcup \mathcal{Y^*}$ (joint scenario). Note that, whatever the scenario, we consider that the function $\phi$ is fixed as well as the pre-trained weights of the seen classes $w_k, \forall k \in \{1,\dots,K\}$.

\subsection{Classifying new classes via NSL}
\label{sec:NSL}

Given that the function $\phi$ and the weights $w_k$ of the seen classes are fixed, our objective is reduced to optimizing the weights $w^*_k, \forall k \in \{1,\dots,K^*\}$ of the unseen classes. Using the cross-entropy as the objective function, this can be expressed as:
\begin{equation*}
\argmin_{w_{1}^*,\dots,w_{K^*}^*} \sum_{i=1}^{n^*} -log (\hat{\eta}_{y^*_i}(x^*_i))
\end{equation*}
\begin{equation*}
\argmin_{w_{1}^*,\dots,w_{K^*}^*} \sum_{i=1}^{n^*} -log\frac{e^{w^*_{y^*_i} \phi(x^*_i)}}{\sum_{j=1}^K e^{w_j \phi(x^*_i)}+\sum_{j=1}^{K{^*}} e^{w^*_j \phi(x^*_i)}}
\label{eq:optim_unseen}
\end{equation*}
In the particular case where we have only one new class (i.e. $K^*=1$), this simplifies to:
\begin{equation*}
\argmax_{w_{1}^*} \sum_{i=1}^{n^*} w_{1}^* \phi(x^*_i)= \argmax_{w_{1}^*} w_{1}^*  \sum_{i=1}^{n^*} \phi(x^*_i)
\end{equation*}
which leads, with the constraints of equation \ref{equ:constraints}, to:
\begin{equation}
w_{1}^* = \frac{1}{n^*}\sum_{i=1}^{n^*} \frac{\phi(x^*_i)}{\|\phi(x^*_i)\|}=\frac{1}{S.n^*}\sum_{i=1}^{n^*} \phi(x^*_i)
\label{eq:Inference}
\end{equation}
The weight $w_{1}^*$ of a new class can thus simply be computed by averaging the feature vectors of the images $x^*_i$ of the new class. This simple theoretical result does not hold anymore when there is more than one novel classes (i.e. when $K^*>1$). However, as we we will see in our experiments, using this estimation procedure for more new classes provides a good approximation of the exact optimal weights and is quite effective in practice. More formally, we propose to estimate the weights $w^*_{k}$ of each of $K^*$ new classes as:
\begin{equation}
\label{equ:average}
w_{k}^* = \frac{1}{S} \frac{\sum_{i=1}^{n^*} \phi(x^*_i)\mathds{1}(y^*_i=K+k)}{\sum_{i=1}^{n^*}\mathds{1}(y^*_i=K+k)} 
\end{equation}

\begin{figure}[!ht]
    \centering
    \includegraphics[width=0.4\textwidth]{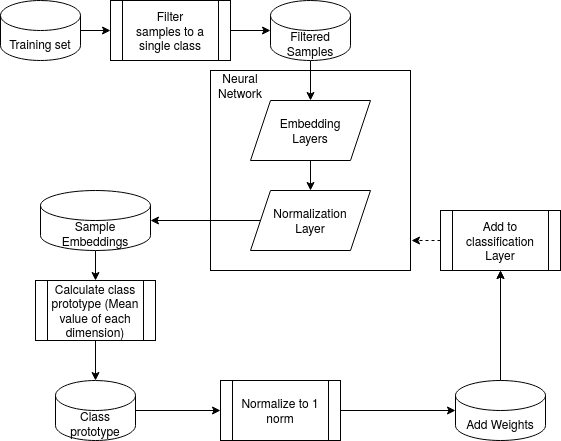}
    \caption{Diagram presenting the approach to infer weights for the decision layer for new classes}
    \label{fig:Diagrama}
\end{figure}

In the \textit{joint scenario}, we are interested in a classifier on both the seen classes and the new classes. This can be expressed as:
\begin{equation}
h_{joint}(x)=\sum_{i=1}^{i=n^*} \max_{k} \frac{e^{w^*_{k} \phi(x^*_i)}}{\sum_{j=1}^K e^{w_j \phi(x^*_i)}+\sum_{j=1}^{K{^*}} e^{w^*_j \phi(x^*_i)}}
\label{eq:JointScenarioEquation}
\end{equation}
where the $w_j$ and $\phi()$ are pre-trained on the seen classes and the new weights $w_{j}^*$ are computed with Equ. \ref{equ:average}.\\
In the \textit{disjoint scenario}, we are interested in a classifier on the new classes only (in a transfer learning way):
\begin{equation}
h_{disjoint}(x)= \sum_{i=1}^{i=n^*}\max_{k} \frac{e^{w^*_{k} \phi(x^*_i)}}{\sum_{j=1}^{K{^*}} e^{w^*_j \phi(x^*_i)}}
\label{eq:DisjointScenarioEquation}
\end{equation}
where $\phi()$ is pre-trained on the seen classes and the new weights $w_{j}^*$ are computed with Equ. \ref{equ:average}.\\
A dataflow depicting our approach to infer the weights for novel classes is presented in Figure \ref{fig:Diagrama}.\\

\iffalse
\begin{figure}[!h]
    \centering
 %   \includegraphics[width=0.4\textwidth]{figs/PlantnetNSL.png}
    \includegraphics[width=0.4\textwidth]{figs/PlantnetNSL.png}
    \caption{Embedding obtained on a subset of the Plantnet training set discussed in section Case Study: The Pl@ntnet dataset when using a latent space with two dimensions using NSL and optimizing until minimum loss}
    \label{fig:PlantNetEmb2D}
\end{figure}
\fi
\iffalse
\begin{figure}[!ht]
    \centering
    %\includegraphics[width=0.5\textwidth]{figs/Cifar100RelativeError.pdf}
    \includegraphics[width=0.4\textwidth]{figs/Cifar100RelativeError.pdf}
    \caption{Relative error for rebuilding the optimized weights by using equation \ref{equ:average} on the cifar 100 dataset when optimizing on a different number of classes on 30 different runs }
    \label{fig:Cifar100Error}
\end{figure}
\fi
%A further discussion of our proposed approach when adding multiple classes to improve our results further will be discussed in section \ref{sec:IncrementalLearning}

\section{Results}
\label{sec:results}
\subsection{Experimental setup}
In this section we present the experimental setup. All experiments presented for the FASHION MNIST and CIFAR datasets were performed using google collaboratory. Experiments using the Pl@ntNet dataset were performed using a Dell PowerEdge R730 server, with 2 CPUs Intel (R) Xeon (R) CPU E5-2690 v3 @ 2.60GHz; 768 GB of RAM; and running on a Linux CentOS 7.7.1908 kernel version 3.10.0-1062.4.3.e17.x86\_64. The machine is equipped with a single NVIDIA Pascal P100 GPU, with 16GB RAM. Implementations were performed using Python 3.7 along with the Keras deep learning library.
\subsection{Evaluating the disjoint scenario}

 In this section we show the results of evaluating the model accuracy on the test set from seen and unseen classes by employing a VGG based model with two blocks.
 
  We optimize the model on $K=10-K^*$ seen classes, and use the trained network to classify the $K^*$ unseen classes without considering the seen ones as possible answers. The approach to do so was presented in equation \ref{eq:DisjointScenarioEquation} in section Proposed Methodology. Results are presented in Tables \ref{tab:DisjointedCifar} and \ref{tab:DisjointedFashion}.
\begin{table}[!ht]
\begin{tabular}{llll}
$K^*$ & NSL &Triplet & Constrative \\
2 & \textbf{0.852} & 0.663 &  0.700\\
3 & \textbf{0.725} & 0.570 & 0.551 \\
4 & \textbf{0.629} & 0.406 & 0.422 \\
5 & \textbf{0.545} & 0.296 & 0.328
\end{tabular}
%\caption{Model results for the cifar10 dataset. $K^*$  refers to the amount of unseen classes. NSL shows model accuracy on unseen classes when a model is optimized on $10-K^*$ classes and only identifies $K^*$. Triplet and constrative refers to the same problem using triplet and constrative loss, respectively and using an K nearest neighboors algorithm trained on the model embedding to perform classification.}
\caption{Model results for the cifar 10 dataset. $K^*$ refers to the amount of unseen classes while other columns refer to the method and accuracy obtained on the test set.}
\label{tab:DisjointedCifar}
\end{table}
%Verificar a triplet
\begin{table}[!ht]
\begin{tabular}{llll}
N & NSL &Triplet & Constrative \\
2 & \textbf{0.897} & 0.62 &  0.703\\
3 & \textbf{0.876} & 0.39 & 0.469 \\
4 & \textbf{0.841} & 0.27 & 0.312 \\
5 & \textbf{0.807} & 0.22 & 0.2
\end{tabular}
\caption{Model results for the Fashion Mnist dataset. $K^*$ refers to the amount of unseen classes while other columns refer to the method and accuracy obtained on the test set.}
\label{tab:DisjointedFashion}
\end{table}

In Tables \ref{tab:DisjointedCifar} and \ref{tab:DisjointedFashion},
 we compare NSL against two metric learning strategies in a disjoint settings. In the first line, we present a scenario in which we train with 8 random classes and evaluate on the other two. The second line trained with 7 and so forth. Our results show that in both datasets the NSL outperformed those metric learning strategies for evaluating novel classes in a disjoint scenario.
 To evaluate both the triplet loss as well as the constrative loss methods we first built the embedding representation, and then feed this representation on an k nearest neighbours model trained on the average embedding of the class using the same number of samples as NSL.
 
\subsection{Evaluating the joint scenario}
In this section we present results when the novel classes must be integrated into the classification process along with the classes used for optimization.
To this end the function that we want to optimize is described in equation\ref{eq:JointScenarioEquation}.
The model is optimized with $10-K^*$ classes and we evaluate the accuracy on these and on the $K^*$ unseen classes, in a 10 possible classes scenario. Results are presented in Tables \ref{tab:JointedCifar} and \ref{tab:JointedFashionMnist}
\iffalse
\begin{table}[!ht]
\begin{tabular}{l|lll|lll|}
& \multicolumn{3}{c}{Seen} & \multicolumn{3}{c}{Unseen} \\
\hline
K^* & NSL &NSL &Triplet &Triplet &Constrative & Constrative \\
2 & 0.559 & 0.501 &0.195 &0.226 &0.205 &0.189 \\
3 & 0.578 & 0.433 &0.128 &0.156  & 0.182 & 0.176  \\
4 & 0.620 & 0.391 &0.0822 &0.127    & 0.166 &0.160 \\
5 & 0.641 & 0.357 &0.0538 &0.146
\end{tabular}
\caption{Model results for the Cifar10 dataset. $K^*$ refers to the amount of unseen classes while other columns refer to the method and accuracy obtained on the test set. Seen shows accuracy on}
\label{tab:JointedCifar}
\end{table}
\fi
\begin{table}[!ht]
\begin{tabular}{l|lll|lll|}
& \multicolumn{3}{c}{Seen} & \multicolumn{3}{c}{Unseen} \\
\hline
$K^*$ & NSL &Triplet &Constrative &NSL &Triplet & Constrative \\
2 & $\mathbf{0.559}$ & 0.195 &0.205 &$\mathbf{0.501}$ &0.226 &0.189 \\
3 & $\mathbf{0.578}$ & 0.128 &0.182 &$\mathbf{0.433}$  & 0.156 & 0.176  \\
4 & $\mathbf{0.620}$ & 0.082 &0.166 &$\mathbf{0.391}$    & 0.127 &0.160 \\
5 & $\mathbf{0.641}$ & 0.05 &0.215 &$\mathbf{0.357}$ &0.146 & 0.176 \\
\end{tabular}
\caption{Model results for the Cifar10 dataset. $K^*$ refers to the amount of unseen classes while other columns refer to the method and accuracy obtained on the test set. Seen refers to the accuracy in the $10-K^*$ classes while unseen on the $K^*$ classes}
\label{tab:JointedCifar}
\end{table}

\begin{table}[!ht]
\begin{tabular}{l|lll|lll|}
& \multicolumn{3}{c}{Seen} & \multicolumn{3}{c}{Unseen} \\
\hline
$K^*$ & NSL &Triplet &Constrative &NSL &Triplet & Constrative \\
2 & $\mathbf{0.854}$ & 0.708 &0.657 &$\mathbf{0.773}$ &0.724 &0.613 \\
3 & $\mathbf{0.842}$ & 0.705 &0.440 &$\mathbf{0.771}$  & 0.716 & 0.390  \\
4 & $\mathbf{0.860}$ & 0.688 &0.284 &$\mathbf{0.739}$    & 0.680 &0.286 \\
5 & $\mathbf{0.878}$ & 0.690 &0.190 &$\mathbf{0.719}$ &0.691 & 0.181 \\
\end{tabular}
 \caption{Model results for the fashion mnist dataset. $K^*$ refers to the amount of unseen classes while other columns refer to the method and accuracy obtained on the test set. Seen refers to the accuracy in the $10-K^*$ classes while unseen on the $K^*$ classes}
\label{tab:JointedFashionMnist}
\end{table}

Tables \ref{tab:JointedCifar} and \ref{tab:JointedFashionMnist} depicts the results of comparing our approach using $NSL$ with metric learning strategies: Triplet loss and constrative loss on both cifar and fashion mnist datasets, considering a joint scenario. The NSL 
outperformed both approaches on the two evaluated datasets, including seen and unseen classes predictions.

\subsection{Case Study: The Pl@ntnet dataset}
\label{sec:plantnet}
\begin{figure}[!ht]
    \centering
    \includegraphics[width=0.3\textwidth]{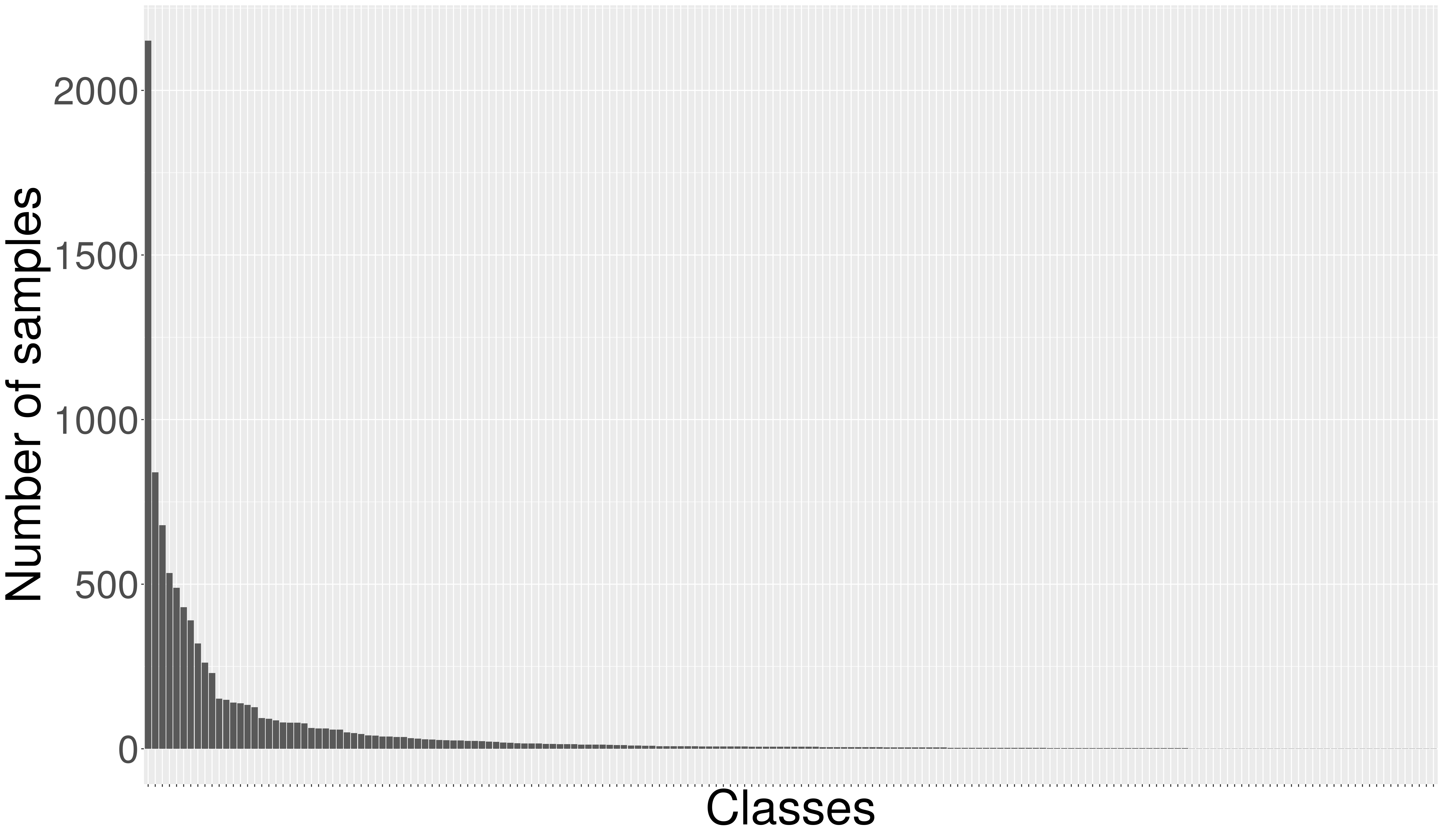}
    \caption{Distribution of the training dataset. Note the long tail distribution presenting how there are many classes with small amounts of data and few with a large amount.}
    \label{fig:PlantNetDistribution}
\end{figure}

    In order to assess on real world data, we evaluate the approach for the closed and open set problems on a dataset built from \textit{Pl@ntnet} database. Pl@ntnet is one of the largest citizen science observatory in the world relying on a mobile application \cite{affouard2017pl} that allows contributors to identify plants using their smartphone (based on convolutional neural networks). The task is challenging as available pictures have different levels of quality, as well as  multiple species from many different parts of the world as shown in figure \ref{fig:PlantNetDistribution}. Given this we wish to evaluate% how our approach fares in the closed set compared to the traditional cross-entropy approach, as well as evaluate 
    it as an open set problem. The problem becomes relevant to the evaluation of the proposed approach given that the scenario is usually represented by a long tail distribution, in which some classes are very common, while others are rare and lack significant available training data.

The subset of the Pl@ntnet data we used was obtained from \cite{garcin} and has a total of 182 classes.

\subsubsection{Experimental design}

As it is clear from Figure \ref{fig:PlantNetDistribution}, there is a high imbalance among classes in the Plantnet dataset. Thus, there are many classes in the training set which have very small amounts of data. Since many plant species have few samples, we are interested in exploring the performance of NSL, where a model is optimized only on more common species and weights to classify uncommon species are inferred, as discussed in section Classifying new classes via NSL. To this end, we perform experiments by optimizing the model only on classes where the number of samples is larger or equal to $N=\{200,100,50,25\}$, which results in $K=\{10,16,28,43\}$, where $K$ is the number of classes, and present the results for joint and disjoint settings. Unseen classes will be selected randomly among those with number of samples $M$, $M<N$, and results will be presented with the average of 30 runs. All models are optimized on 100 epochs and weights that minimize validation loss are used for inference. The model architecture is four convolutional blocks with a 3x3 kernel, the first two with 64 filters and the last two with 256 filters followed by a 2x2 MaxPooling layer, After we define an convolutional block as two convolutions followed by an Maxpooling layer, we implement two convolution blocks with (256,512) and (512,1024) filters followed by a flatten layer and a dense layer with 1024 units with no activation, this layer output is normalized to $S$ norm and fed to the classifier. Pre-processing steps on the data only include normalizing it to $[0,1]$ range and reshaping it to a $<96,96,3>$ shape. Models are optimized with weighted cross-entropy by passing the class weights arguments to keras fit function to take the class imbalance into account on the loss functions, ensuring that solutions that output only the majority class are penalized. For the open set tasks, we report balanced accuracy to better take into account class imbalance.

%\begin{figure}[!ht]
 %   \centering
 %   \includegraphics[width=0.2\textwidth]{NetworkArchitecturePlantnet.png}
 %   \caption{Network architecture used to perform the Plantnet experiments}
 %   \label{fig:Archictecture}
%\end{figure}

\subsubsection{Results}
In this section we present results on the model balanced accuracy for both the disjoint and joint open set problems when the model was optimized on different number of classes. As already discussed, we selected seen classes based on a filter on the number of samples $\geq N$. Given this, the relation between the number of samples and the number of classes is as follows: $[25,43],[50,28],[100,16],[200,10]$.

\begin{table}[]
    \centering
    \begin{tabular}{c|c|c|c|c|c}
        Number of classes & 10&16&28&43 \\ \hline
        NSL accuracy & 0.7349&0.6617&0.5382&0.3974\\
    \end{tabular}
    \caption{Model accuracy on the test set optimized for 100 epochs on weighted cross-entropy}
    \label{tab:NSLvsSLPlantnet}
\end{table}
\iffalse
\begin{table}[]
    \centering
    \begin{tabular}{c|c}
       Number of classes  &  NSL accuracy  \\
        10     & \textbf{0.7349} \\
        16 & \textbf{0.6617}\\
        28 &\textbf{0.5382}\\
        43 & \textbf{0.3974}\\
    \end{tabular}
    \caption{Model accuracy on the test set optimized for 100 epochs on weighted cross-entropy}
    \label{tab:NSLvsSLPlantnet}
\end{table}
\fi

In table \ref{tab:NSLvsSLPlantnet} we present the model accuracy for the four different models that will be used in the plantnet analysis. All models were optimized by receiving the same amount of samples per epoch as well as number of epochs. These set of models will then be used to evaluate both the disjoint and joint scenarios in the further sections.
\subsubsection{Disjoint scenario}
\begin{figure}[!ht]
    \centering
    \includegraphics[width=0.4 \textwidth]{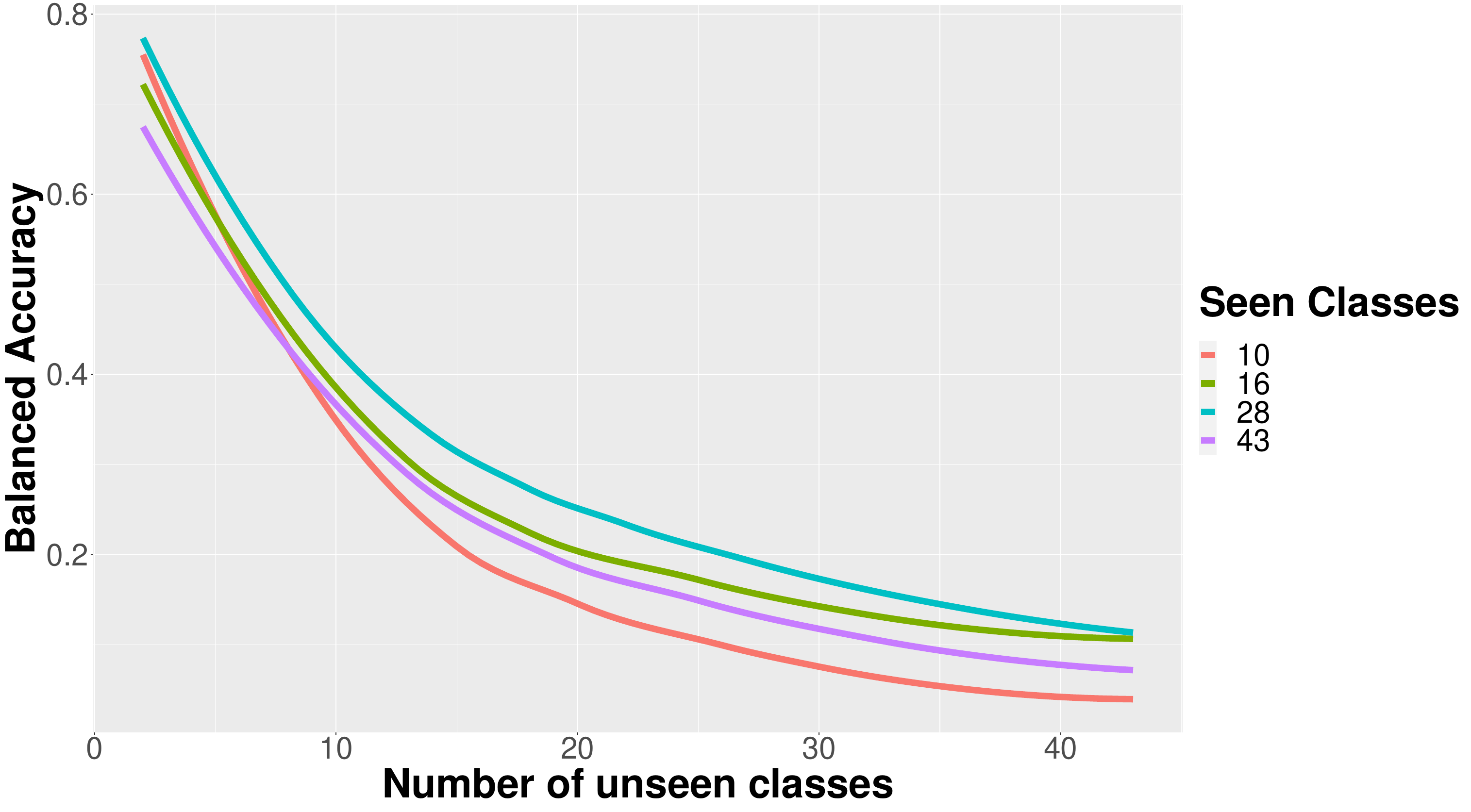}
    \caption{Comparing models with the same architecture and optimized upon the same amount of data and number of epochs, but with different amount of seen classes on their ability to classify novel classes}
    \label{fig:Comparative}
\end{figure}

In this subsection we present the disjoint analysis for the Plantnet dataset. We instantiate our base model without the last layer and then perform classification on novel classes randomly sampled from the total unseen classes by inferring the weights between the penultimate layer and the decision layer. We present balanced accuracy on the model capacity on novel classes by performing 30 runs for each value of $K^*$. Results are presented in figure \ref{fig:Comparative}.
\iffalse
\begin{figure}[!ht]
    \centering
    \includegraphics[width=0.4\textwidth]{figs/TrainedOn200Points.pdf}
   % \includegraphics[width=0.5\textwidth]{figs/DisjointedBalanced10.pdf}
    \caption{Model results when classifying among classes that were unseen on the training set by using inferred weights and optimized on 10 classes}
    \label{fig:16classes}
\end{figure}

\begin{figure}[!ht]
    \centering
    \includegraphics[width=0.4\textwidth]{figs/TrainedOn100Points.pdf}
    %\includegraphics[width=0.5\textwidth]{figs/DisjointedBalanced16.pdf}
    \caption{Model results when classifying among classes that were unseen on the training set by using inferred weights and optimized on 16 classes}
    \label{fig:16classes}
\end{figure}

\begin{figure}[!ht]
    \centering
    \includegraphics[width=0.4\textwidth]{figs/TrainedOn50Points.pdf}
    %\includegraphics[width=0.5\textwidth]{figs/DisjointedBalanced28.pdf}
    \caption{Model results when classifying among classes that were unseen on the training set by using inferred weights and optimized on 28 classes}
    \label{fig:28classes}
\end{figure}

\begin{figure}[!ht]
    \centering
    \includegraphics[width=0.4\textwidth]{figs/TrainedOn25Points.pdf}
    %\includegraphics[width=0.5\textwidth]{figs/DisjointedBalanced43.pdf}
    \caption{Model results when classifying among classes that were unseen on the training set by using inferred weights and optimized on 43 classes}
    \label{fig:43classes}
\end{figure}

\fi

In figure \ref{fig:Comparative}  we show a comparison between models trained on a different number of classes with the same amount of data by presenting their ability to identify novel classes in a disjoint scenario. Our results show that diversity of classes seen during training allowed the model to become more robust for novel classes as the model trained with 10 classes performed the worst of all for novel classes. However it's also important to note that optimizing on a higher number of classes is a more complex problem, requiring more data to be seen by the model, more updates or a more complex model to learn robust features for all seen classes. This is shown on the curve obtained from the model optimized upon 43 classes, which ranks the second worst. Our best result was seen on the model with 28 classes, which had a high class diversity while also learning robust features during training. 

\subsubsection{Joint scenario}
\begin{figure}[!ht]
    \centering
    \includegraphics[width=0.4\textwidth]{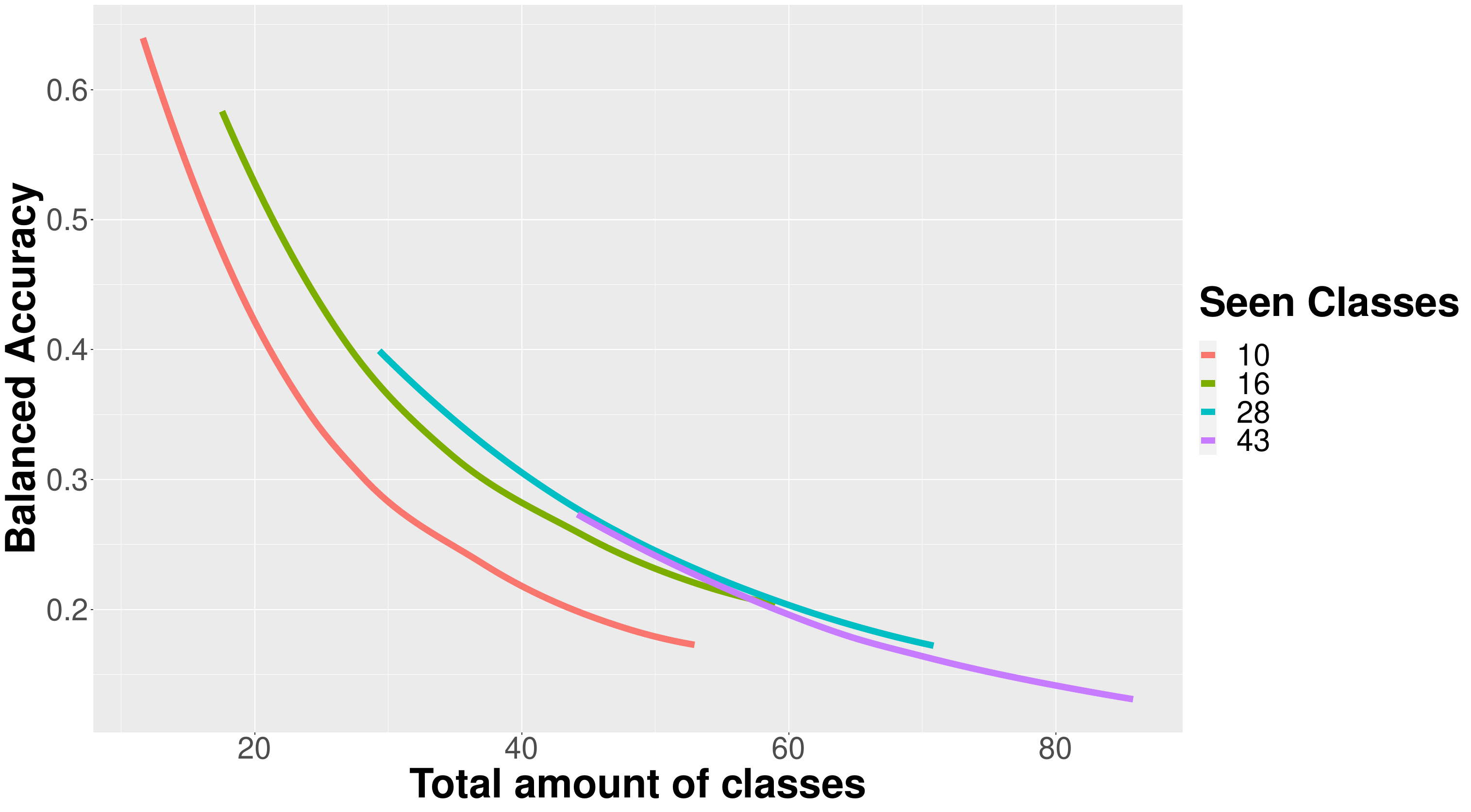}
    \caption{Analysis for the joint scenario showing the results for different models on overall class architecture }
    \label{fig:JointedOverAll}
\end{figure}

\begin{figure}[!ht]
    \centering
        \includegraphics[width=0.4\textwidth]{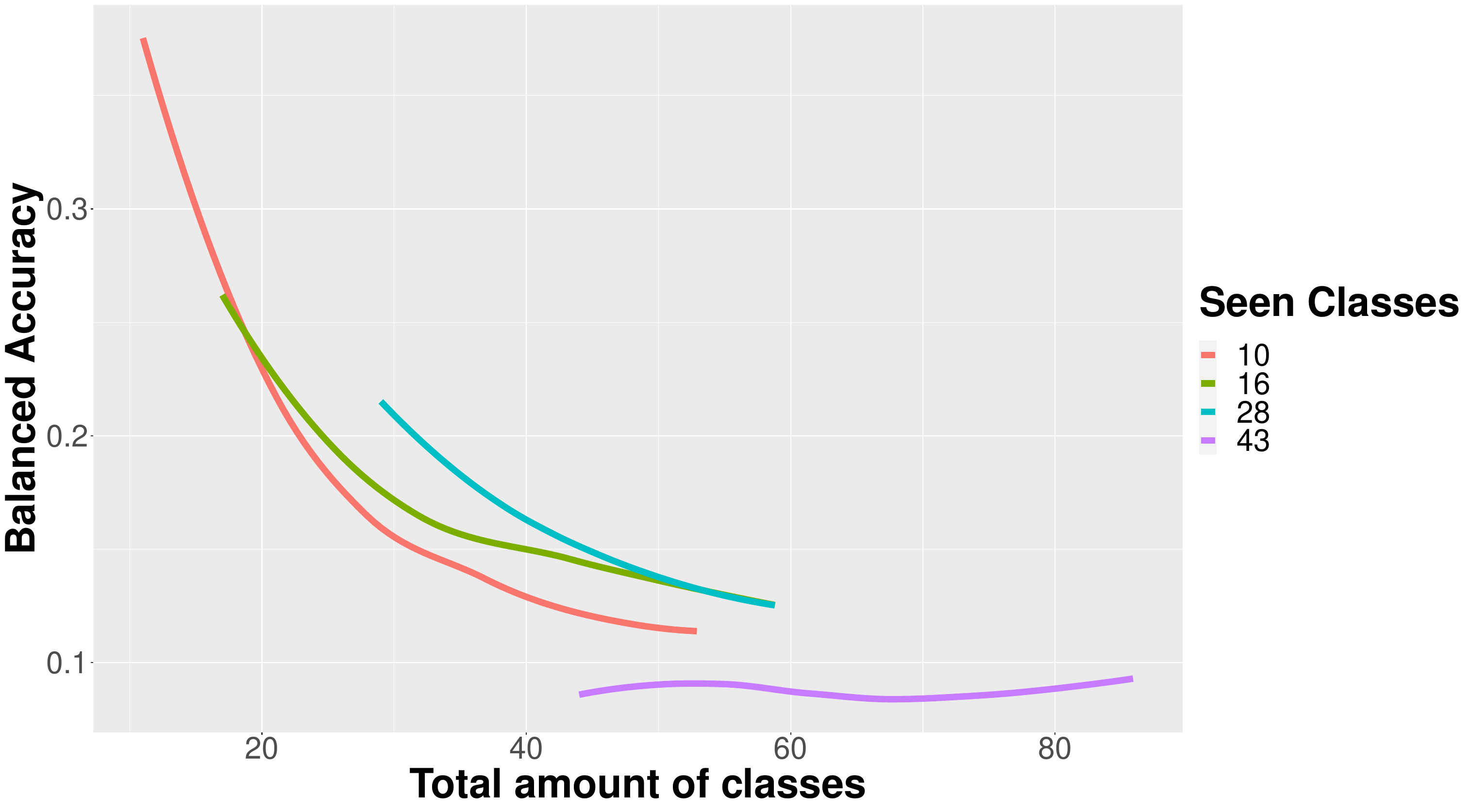}
    \caption{Analysis for the joint scenario showing the results for different models on the unseen classes }
    \label{fig:JointedOverAllUnseen}
\end{figure}

In this subsection, we present the results of the joint scenario for the Pl@ntnet dataset. We present results where we instantiate the base model trained on $K$ classes. Then, we add other $K^*$ unseen classes so that the model must classify between $K+K^*$ classes. Weights for the $K^*$ classes are inferred as described in section Classifying new classes via NSL and we report model accuracy for the overall model, as well as for the $M$ classes. Results are presented in figures \ref{fig:JointedOverAll} and \ref{fig:JointedOverAllUnseen}.
\iffalse
\begin{figure}[!ht]
    \centering
    \includegraphics[width=0.4\textwidth]{figs/JointedOn10Classes.pdf}
    \caption{Analysis for the joint scenario where seen and unseen classes are possible answers for the model. Results show a model trained  on 10 classes.}
    \label{fig:Jointed10}
\end{figure}

\begin{figure}[!ht]
    \centering
    \includegraphics[width=0.4\textwidth]{figs/JointedOn16Classes.pdf}
    \caption{Analysis for the joint scenario where seen and unseen classes are possible answers for the model. Results show a model trained  on 16 classes.}
    \label{fig:Jointed16}
\end{figure}

\begin{figure}[!ht]
    \centering
    \includegraphics[width=0.4\textwidth]{figs/JointedOn28Classes.pdf}
    \caption{Analysis for the joint scenario where seen and unseen classes are possible answers for the model. Results show a model trained  on 28 classes.}
    \label{fig:Jointed28}
\end{figure}

\begin{figure}[!ht]
    \centering
    \includegraphics[width=0.4\textwidth]{figs/JointedOn43Classes.pdf}
    \caption{Analysis for the joint scenario where seen and unseen classes are possible answers for the model, Results show a model trained  on 43 classes}
    \label{fig:Jointed43}
\end{figure}

\fi

In figures \ref{fig:JointedOverAll}  and \ref{fig:JointedOverAllUnseen} we present our results for the joint scenario evaluating the overall model quality as new classes are added as well as showing the balanced accuracy calculated only on the unseen classes. Our conclusions on the disjoint scenario seen in figure \ref{fig:Comparative} also hold in these results, as we can see for both of these cases the model trained on 28 classes has the higher balanced accuracy given the same number of total classes and the model optimized on 43 classes shows the worse balanced accuracy. The lack of diversity of the model optimized on 10 classes can be seen influencing it's quality as the number of novel classes increase in both scenarios.

\subsection{Few shot scenario for incremental learning}
\label{sec:IncrementalLearning}
In Equation \ref{eq:Inference}, we show how our proposed methodology of inferring weights actually finds the set of weights that minimizes cross-entropy,  whenever a single novel class is included. However, when including multiple classes, our proposal may not yield the optimum set of weights for each new neuron. In this section we present a set of experiments comparing the performances obtained by our inferred weights with the ones obtained through incremental learning, i.e. by minimizing the cross-entropy loss on samples of the new classes while freezing all the other network weights. Experiments were performed using the Cifar10 dataset. The initial model is  trained on the training samples of $K$ seen classes and the incremental learning phase is computed on the training samples of $K^*$ unseen classes (while freezing all the other network weights).

To evaluate the proposed strategy in a few shot scenario, we train the Resnet50 architecture using the NSL constraint, as provided by keras, on a subset of classes of the cifar10 dataset. Once the network has been trained for 100 epochs on a subset of the dataset, we sample a small number of examples of the classes that were unseen during training in a few shot scenario (one,  five and twenty five shots). We infer the weights for the novel classes using the methodology described in section Classifying new classes via NSL and measure the model quality on the test set compared to the incremental learning approach (using the same few shots for each class).

\begin{figure}[!ht]
    \centering
    \includegraphics[width=0.4\textwidth]{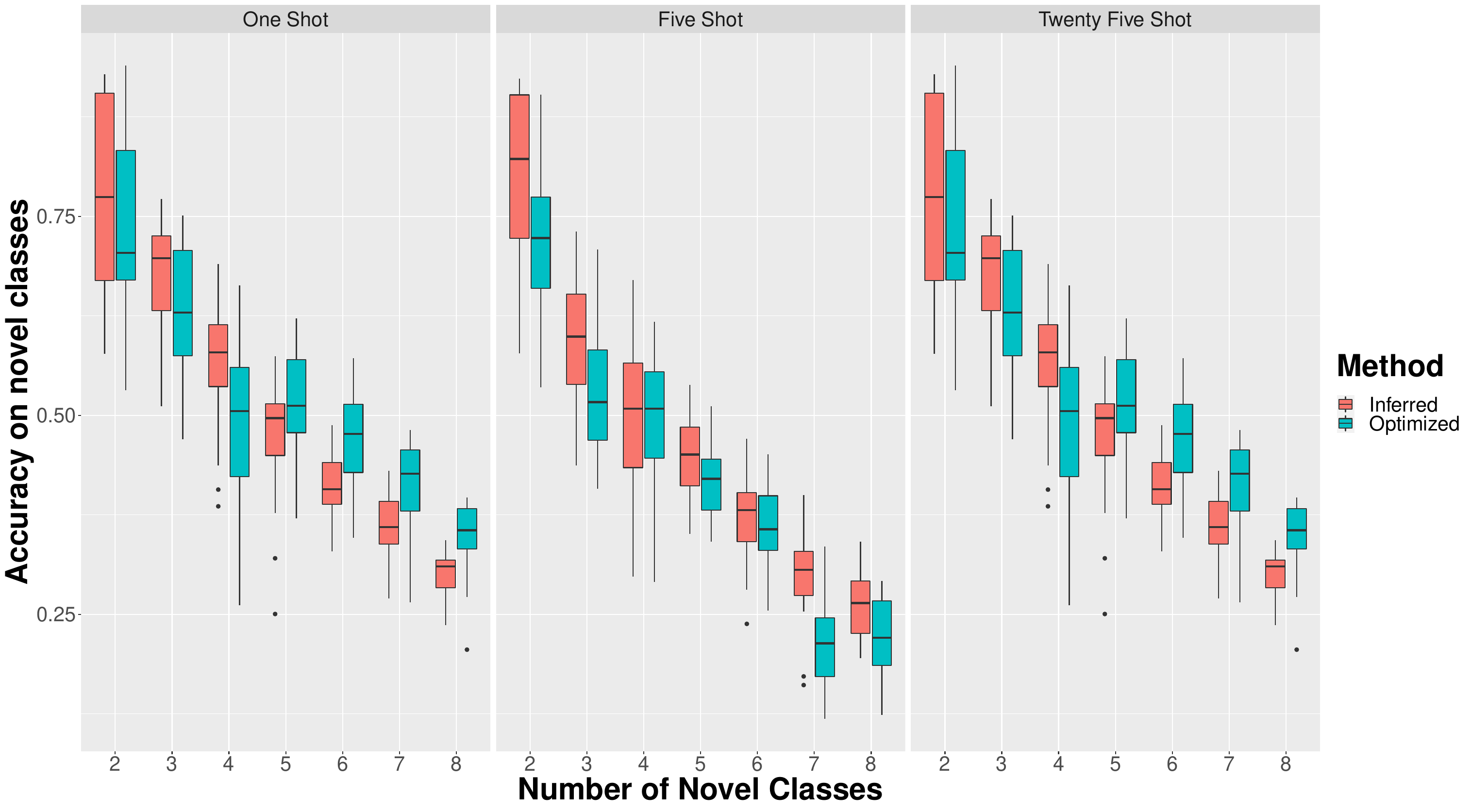}
        \caption{Accuracy obtained in different few shot scenarios for the cifar10 dataset when comparing using inferred weights versus further optimizing them. Note that when the number of novel classes is smaller or equal to the number of seen classes the model that does not use further optimization tends to have bigger accuracy }
    \label{fig:Cifar10ResNetAccuracyFewShot}
\end{figure}

%After we optimize the inferred weights in respect to the same few samples the weights were inferred on, while freezing all other weights for 10 epochs and apply the model again on the test set, the same strategy that is already described in section Few shot scenario for incremental learning but for a few shot scenario. 
%Once the inferred weights have been optimized using the same few samples,  we apply the model on the same test set, which corresponds to the strategy described in \ref{sec:IncrementalLearning} but considering a few shot scenario.
We report the results for one, five and twenty five shot scenarios on the CIFAR10 dataset.% Figure \ref{fig:Cifar10ResNetLossFewShot} presents boxplots of the cross-entropy loss (computed on the test set). We observe a competitive loss exhibited by the inferred weights approach. 
Figure \ref{fig:Cifar10ResNetAccuracyFewShot} depicts similar results for all scenarios, considering the accuracy of predictions.  

%, as well as model accuracy, using inferred weights in a few shot scenario.   on the novel classes which is presented in figure \ref{fig:Cifar10ResNetAccuracyFewShot} which presents how the model accuracy was improved in the few shot scenario when further optimization is not performed.
\iffalse
\begin{figure}[!ht]
    \centering
    \includegraphics[width=0.4\textwidth]{ResnetIncremental/CrossEntropyLossCifar.pdf}
    \caption{cross-entropy Loss obtained in different few shot scenarios for the cifar10 dataset when comparing using inferred weights versus further optimizing them, note that when the number of novel classes is smaller or equal to the number of seen classes the model that does not use further optimization tends to have smaller loss }
    \label{fig:Cifar10ResNetLossFewShot}
\end{figure}
\fi

%\fi

\section{Related Work}
\label{sec:Related}
Creating models that are able to classify novel classes is a task that is explored in different fields of artificial intelligence. There are two fields our work falls upon. The first is called metric learning discussed in section Classifying new classes, while the second is called incremental learning briefly presented in section Few shot scenario for incremental learning.

Metric learning which is a sub-field of few shot learning is a field where one aims to train a model to identify classes via some property in a metric space enforced during training. The enforced property can be for example that examples of the same class form a cluster according to some predefined metric and each class has its own cluster. These properties can then be explored to identify novel classes given that it is possible to determine the cluster of a novel class with some labeled examples without retraining the model. There are many works that fall in this category \cite{TripletLossPaper},\cite{ConstrativeLossPaper},\cite{constelationloss},\cite{MatchingNetworkPaper}. While all these approaches enforce metric properties on the latent space, they also require pairwise training, which our approach does not require.
\iffalse
The second field our work falls upon is incremental learning. In incremental learning one aims to add novel classes to a model by minimizing the loss in relation to the examples of the novel classes with an added constraint that all weights that were previously trained should stay fixed. Some interesting works in the field are \cite{ArtigoIncLearningAlexis},\cite{castro2018end-to-end},\cite{Chaudhry_2019_1},\cite{Kirkpatrick3521}.
We discussed in section Few shot scenario for incremental learning how our work can benefit from the strategies in the field when including many novel classes in our approach. Another aspect  that emphasizes the contribution of our approach in the field is raised by \cite{ArtigoIncLearningAlexis}, who argues that a good incremental learning method should be able to optimize on new classes without access to the old training data. To be capable of handling novel classes without prior information, as is the case of zero shot learning. \textit{Model size should remain relatively unchanged after learning new classes}\cite{ArtigoIncLearningAlexis}. We argue that all these points are satisfied since we are able to infer weights for the novel class using equation \ref{eq:Inference} by using only the novel class. We do not require prior knowledge to do so, and to add $t$ novel classes we only need to store a matrix of size $M \times t$ where $M$ is the dimension of the latent space.
\fi

\section{Conclusion}
\label{sec:conclusion1}

In this paper, we presented how the normalized softmax loss can be employed on the open set problem. We presented results on different datasets for both the disjoint and joint open set problems and compared them to metric learning strategies. We show that the NSL based approach demonstrates superior results producing  more robust features and implementing a less costly optimization procedure, as it does not require pairwise training. Results on a real world use case evaluating on a subset of the Pl@ntnet data shows how our approach can be employed to identify classes unseen during optimization, with weights associated to the classification of new data inferred by the approach.% We also presented how our approach can be looked via the lenses of incremental learning, and that in those lenses we provide properties for a good incremental learning model along with being able to start the novel weights close to the optimum when including multiple novel classes.
%\end{document}  % This is where a 'short' article might terminate

% ensure same length columns on last page (might need two sub-sequent latex runs)
\balance

%ACKNOWLEDGMENTS are optional
\section{Acknowledgments}
 The authors would like to thank Petrobras for supporting this work through the project "Development of an Intelligent software platform". We would also like to thank the INRIA-Brazil Associated Team cooperation project HPDaSc.

%\bibliography{aaai22.bib}

\section{Reproducibility Statement}

In this section we detail the steps we ensured to ensure that our work is reproducible. To ensure data availability we mostly use public datasets that are available in the keras.datasets interface. The subset of the Pl@ntnet dataset we used in this paper available as numpy arrays in the plantnet folder that is available via an google drive link that is presented in the appendix section.

Regarding data preprocessing all preprocessing steps as well as the structure of the models are presented in the main paper. 

Concerning the mathematical formulation of the problem. the main mathematical formulation is presented in the main paper while additional information about the area of the problem is presented in the appendix on metric learning and the latent space.

Lastly regarding experiment reproducibility all experiments were organized into jupyter notebooks and these are organized into an folder that is available via an google drive link on the appendix.

\bibliography{iclr2022_conference}
\bibliographystyle{iclr2022_conference}

\appendix
\section{Appendix}
The following sections contain additional analysis and theoretical discussion that is not integral to the understanding of the paper but the authors would like to share about the research performed.

\subsection {Metric Learning and the latent space}
Deep neural networks learn a set of transformations and relations among the inputs in order to obtain the desired output during optimization. The network can be broken into two parts: the projection module ($\phi(x)$), which takes the data and transforms it into a representation in the latent space; and the data representation processing, applied by the layer implementing the desired task ($\psi(z)$). The latter enforces the latent space to have some properties that are defined by the task. 
%For example, the regression task is defined as $y=\sum_iw_ix+b$ and so given a fixed $w$ if $x\approx x'$ then $y\approx y' $.
When the network is trained with a cross-entropy loss, the objective is the following:
\begin{equation*}
\argmin_{\theta} \sum_{i=1}^{n} -log(\hat{\eta}_{y_i}(x_i)) 
\end{equation*}
where $\theta$ is the set of all parameters of the network (for both $\psi(x)$ and $\phi(x)$). Thus, after optimization, we generally have that $\hat{\eta}_{y_i}(x_i)>>\hat{\eta}_{j}(x_i)$ for $j\neq y_i$ which can only be achieved if: $w_{y_i} \phi(x_i) + b_{y_i} >> w_j \phi(x_i) + b_j$ for $j\neq y_i$. In other words, the \textit{softmax cross-entropy} approach enforces the inequality $w_i z_i +b_i >> w_j z_i+b_j$ in the latent space, where $i,j$ represent different classes \cite{NSL}.
%When discussing the classification problem, . 
A proposed alternative that optimizes the latent space directly and can enforce metric properties that allows the model to be used for novel classes is know as \textit{metric learning}.

The metric learning approach learns a set of features that obey a metric distance on the latent space. The model can be optimized to learn a similarity metric between pairs, as proposed in \cite{RelationNetwork}, or can enforce the latent space to obey a predefined metric distance like euclidean distance or cosine similarity. Some strategies, such as the \textit{constrative loss} \cite{ConstrativeLossPaper}, learn on pairs of data, while others learn using triplets like \textit{triplet loss} \cite{TripletLossPaper}. 

Optimization on these approaches aim to obtain disjoint clusters for each class of interest in the latent space, according to a predefined metric distance. As a desired consequence of the approach, classification can be performed for novel classes by using the representation of an anchor example and calculating the metric distance between a query point and the anchor.

\subsection{How does the amount of samples affect the class prototype ?}

\begin{figure}[H]
    \centering
    \includegraphics[width=0.4\textwidth]{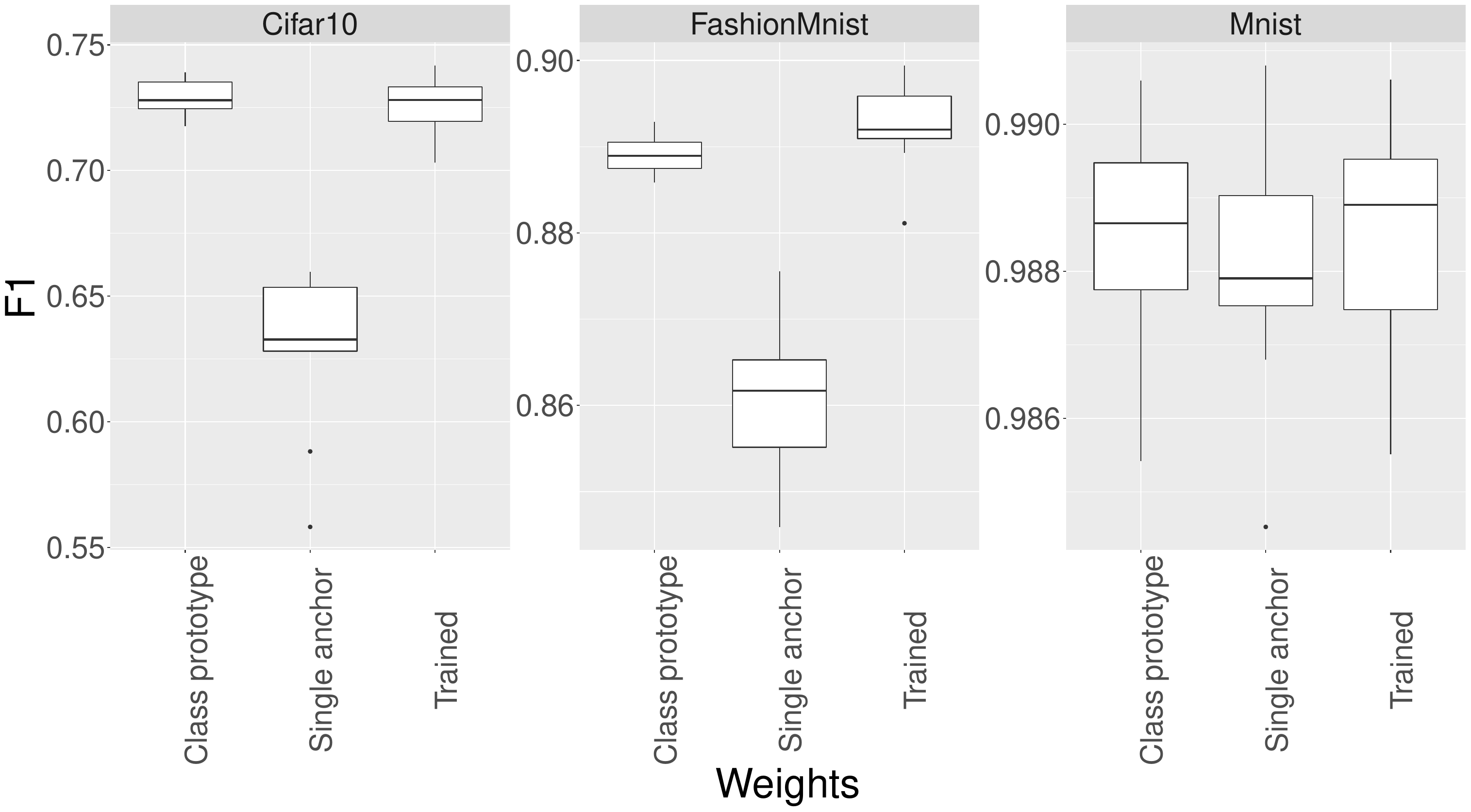}
    \caption{F1 score on the test set for three different datasets where we show results when weights are: (a) Trained: the ones found during optimization; (b) Single anchor: weights are inferred using a single random anchor example; and (c) weights are inferred using the training set to build a class prototype.}
    \label{fig:NValue}
\end{figure}

A common scenario in which the identification of unseen classes appears is the one where the amount of available data samples for classes of interest is small or the cost of optimizing another model to include the new classes becomes too costly. Therefore, all strategies discussed in this paper classify new classes based on labeled examples without retraining. The influence of the number of samples needed to perform classification tasks using the NSL approach is shown in three different datasets in Figure \ref{fig:NValue} for a model optimized for 30 epochs. We use keras base learning rate with adam optimizer. The experiment consider that the model was trained on all ten classes. Models for mnist and fashion mnist consider only dense layers, while cifar considers two convolutional blocks with 32 and 64 filters each. % We define class prototype in the same way as \cite{PrototipicalNetworks}.
We create the class prototype using the inferred weights obtained via Eq \ref{equ:average}, an strategy similar to \cite{PrototipicalNetworks}. 

As it can be seen on Figure \ref{fig:NValue}, weights inferred using Equation \ref{eq:Inference} on the training set maintain  the same model accuracy as by using the weights obtained during optimization. We also can observe that model quality, when inferring via a single example, decays in relation to the task complexity. On the x axis we present how the class weights were obtained where class prototype uses the whole training set to infer the weights according to our methodology, while single anchor uses a single random example from the training set for weight inference. Y axis presents the F1-Score on the test set.

\subsection{Comparison to incremental learning using many samples}
\label{sec:IncrementalLearning}
%Another field in deep learning which aims to integrate novel classes into a model is called incremental learning \cite{ArtigoIncLearningAlexis}. In this field, the objective is to minimize the loss function with the constraints that all the network weights would be frozen except the weights that connect to the new neuron. 
In Equation \ref{eq:Inference}, we show how our proposed methodology of inferring weights actually finds the set of weights that minimizes cross-entropy,  whenever a single novel class is included. However, when including multiple classes, our proposal may not yield the optimum set of weights for each new neuron. In this section we present a set of experiments comparing the performances obtained by our inferred weights with the ones obtained through incremental learning, i.e. by minimizing the cross-entropy loss on samples of the new classes while freezing all the other network weights. Experiments were performed using the fashion mnist dataset. The initial model is  trained on the training samples of $K$ seen classes and the incremental learning phase is computed on the training samples of $K^*$ unseen classes (while freezing all the other network weights).

\begin{figure}[!ht]
    \centering
        \includegraphics[width=0.4\textwidth]{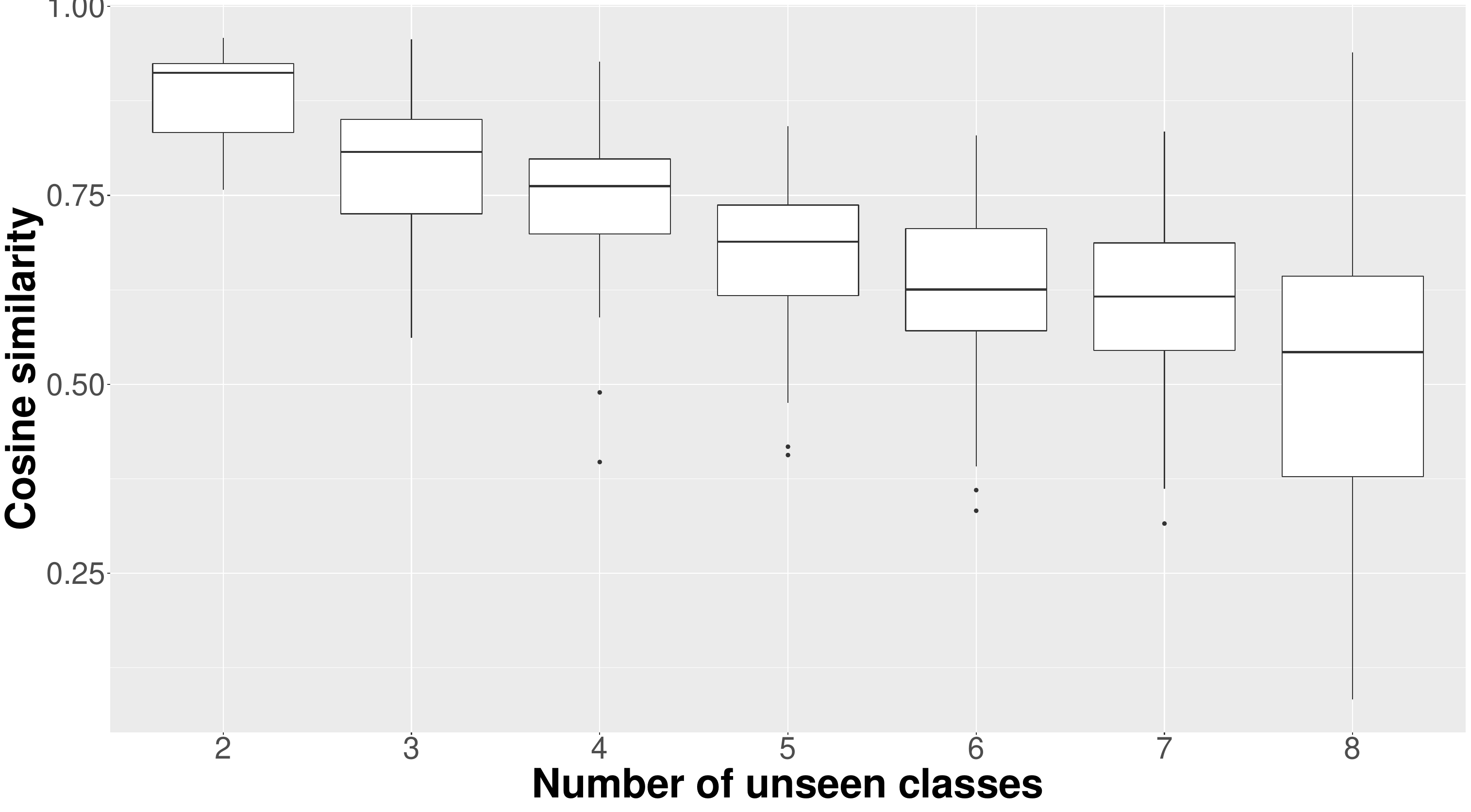}
    \caption{Cosine similarity between inferred weights, obtained as in equation \ref{eq:Inference}, and optimized weights, obtained through cross-entropy optimization using the incremental learning constraint, for different numbers of novel classes on the fashion MNIST dataset }
    \label{fig:Increment1}
\end{figure}

%In figure \ref{fig:Increment1}, we compare the weights associated to an increasing number of unseen classes when computed using cross-entropy optimization and inferred using the proposed approach.
%we present the mean cosine similarity between weights obtained during cross-entropy optimization, given the constraints, and weights inferred via equation \ref{eq:Inference}. 
 Figure \ref{fig:Increment1} first shows the cosine similarity between the inferred and the optimized weights for different numbers of unseen class. As we can observe, when the number of novel classes is small, the two sets of weights are almost identical, which means that the inferred weights are as good as the ones optimized through incremental learning (while being much faster and simpler to compute). With larger numbers of novel classes, we can observe that the mean cosine similarity is still very high. This suggests that the gain of incremental learning might not be very high in this case as well.

\begin{figure}[!ht]
    \centering
    \includegraphics[width=0.4\textwidth]{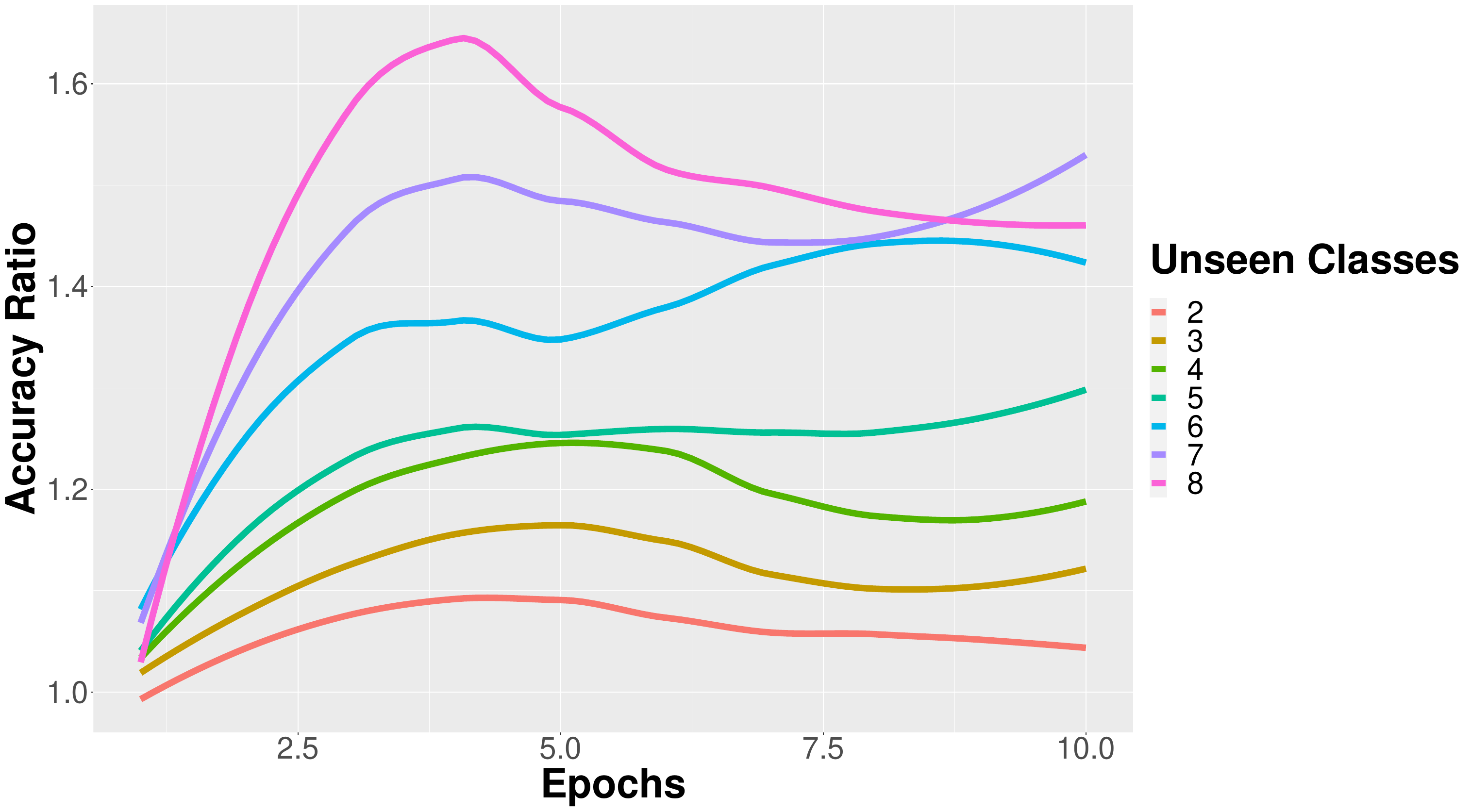}
    \caption{Ratio of accuracy after optimization versus inferred weights on $unseen=10-seen$ classes when optimization occurs on seen classes for the fashion mnist dataset.}
    \label{fig:Increment3}
\end{figure}

%\begin{figure}[!ht]
%    \centering
%    \includegraphics[width=0.4\textwidth]{figs/IntraClassSimilarityInferredVsOptimized.png}
 %   \caption{Intra class similarity between inferred weights obtained in equation \ref{eq:Inference} and optimized weights for different numbers of novel classes on the fashion mnist dataset }
  %  \label{fig:Increment2}
%\end{figure}
To quantify this gain, Figure \ref{fig:Increment3} presents the ratio of the accuracy achieved through incremental learning over the one obtained with the inferred weights (using MNIST test set). We see that when including small number of novel classes the ratio stays close to 1, showing no strong accuracy improvement due to optimization. When including more classes, the gain of incremental learning can be higher (up to 1.55 for 8 unseen classes) but this requires 4-5 epochs on the training set. This suggests that the inferred weights may be used to initialize the incremental phase and get a faster convergence when having a lot of data for the novel class. In the next section we perform the same experiment considering small data.

\begin{figure}[!ht]
    \centering
   \includegraphics[width=0.4\textwidth]{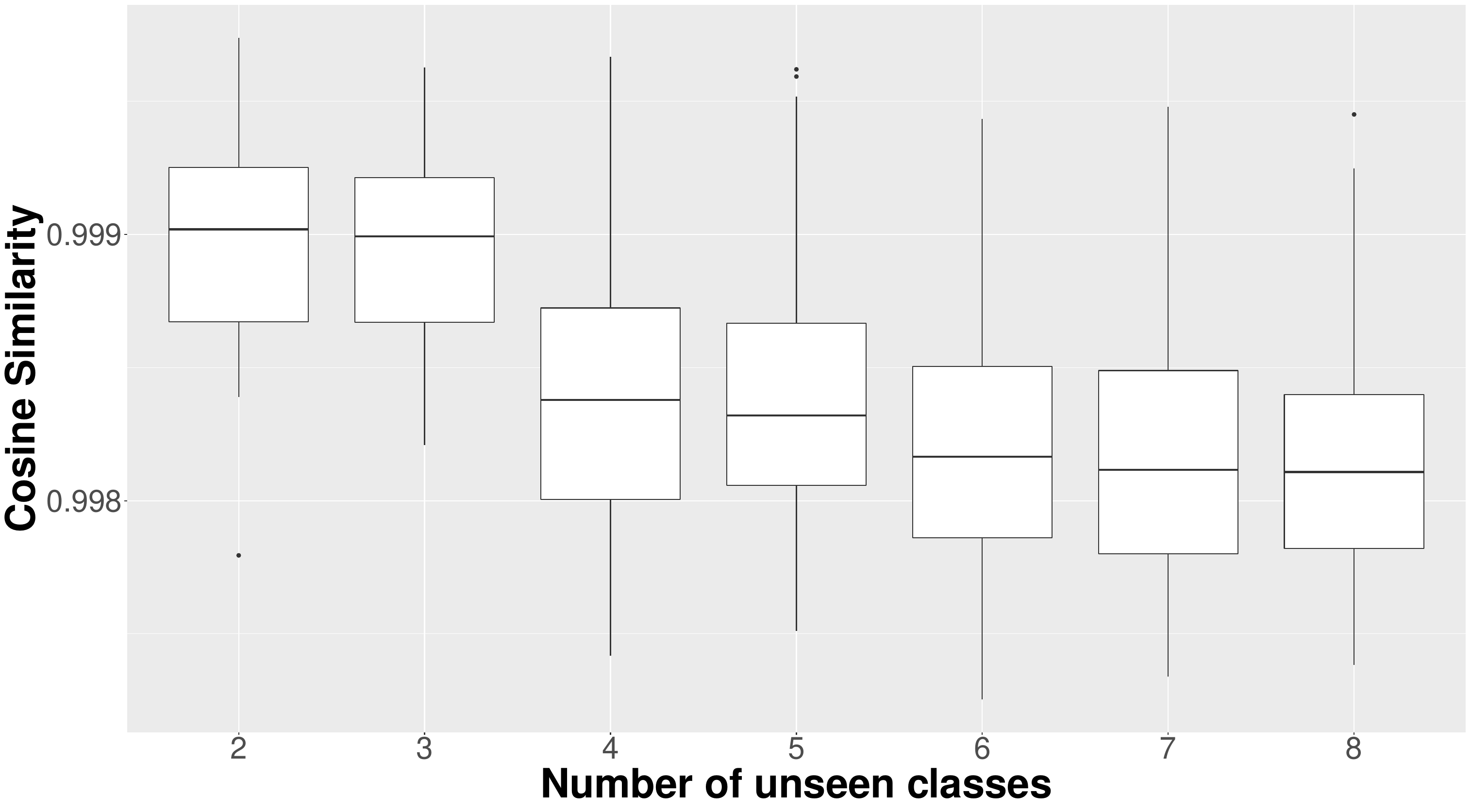}
    \caption{Cosine similarity between inferred weights, obtained as in equation \ref{eq:Inference}, and optimized weights, obtained through cross-entropy optimization using the incremental learning constraint, for different numbers of novel classes on the fashion MNIST dataset when using only two samples to generate the novel class }
    \label{fig:Increment1}
\end{figure}

\section{Experiments}

All performed experiments are available via the following google drive folder: \url{https://drive.google.com/drive/folders/1P2WUw11k9s1IbSdqT38nNbM61m6KQyfD?usp=sharing}

The subset of the plantnet data is already available in numpy array format inside the plantnet folder in this link

\end{document}

%% file: math_commands.tex
\usepackage{amsmath,amsfonts,bm}

\def\eqref#1{equation~\ref{#1}}
\def\1{\bm{1}}

\DeclareMathAlphabet{\mathsfit}{\encodingdefault}{\sfdefault}{m}{sl}
\SetMathAlphabet{\mathsfit}{bold}{\encodingdefault}{\sfdefault}{bx}{n}

\DeclareMathOperator*{\argmax}{arg\,max}
\DeclareMathOperator*{\argmin}{arg\,min}

%% file: iclr2022_conference.bbl
\begin{thebibliography}{14}
\providecommand{\natexlab}[1]{#1}
\providecommand{\url}[1]{\texttt{#1}}
\expandafter\ifx\csname urlstyle\endcsname\relax
  \providecommand{\doi}[1]{doi: #1}\else
  \providecommand{\doi}{doi: \begingroup \urlstyle{rm}\Url}\fi

\bibitem[Affouard et~al.(2017)Affouard, Go{\"e}au, Bonnet, Lombardo, and
  Joly]{affouard2017pl}
Antoine Affouard, Herv{\'e} Go{\"e}au, Pierre Bonnet, Jean-Christophe Lombardo,
  and Alexis Joly.
\newblock Pl@ntnet app in the era of deep learning.
\newblock In \emph{ICLR: International Conference on Learning Representations},
  2017.

\bibitem[Fields(2016)]{HumanVision}
Chris Fields.
\newblock Editorial: How humans recognize objects: Segmentation, categorization
  and individual identification.
\newblock \emph{Frontiers in Psychology}, 7:\penalty0 400, 2016.
\newblock ISSN 1664-1078.
\newblock \doi{10.3389/fpsyg.2016.00400}.
\newblock URL
  \url{https://www.frontiersin.org/article/10.3389/fpsyg.2016.00400}.

\bibitem[Garcin()]{garcin}
Camille Garcin.
\newblock Projects · garcin camille / plantnet\_dataset.
\newblock URL \url{https://gitlab.inria.fr/cgarcin/plantnet_dataset}.

\bibitem[Hadsell et~al.(2006)Hadsell, Chopra, and LeCun]{ConstrativeLossPaper}
Raia Hadsell, Sumit Chopra, and Yann LeCun.
\newblock Dimensionality reduction by learning an invariant mapping.
\newblock In \emph{Proceedings of the 2006 IEEE Computer Society Conference on
  Computer Vision and Pattern Recognition - Volume 2}, CVPR ’06, pp.\
  1735–1742, USA, 2006. IEEE Computer Society.
\newblock ISBN 0769525970.
\newblock \doi{10.1109/CVPR.2006.100}.
\newblock URL \url{https://doi.org/10.1109/CVPR.2006.100}.

\bibitem[{He} et~al.(2016){He}, {Zhang}, {Ren}, and {Sun}]{resnet}
K.~{He}, X.~{Zhang}, S.~{Ren}, and J.~{Sun}.
\newblock Deep residual learning for image recognition.
\newblock In \emph{2016 IEEE Conference on Computer Vision and Pattern
  Recognition (CVPR)}, pp.\  770--778, 2016.
\newblock \doi{10.1109/CVPR.2016.90}.

\bibitem[KAYA \& BİLGE(2019)KAYA and BİLGE]{MetricSurvey}
Mahmut KAYA and Hasan~Şakir BİLGE.
\newblock Deep metric learning: A survey.
\newblock \emph{Symmetry}, 11\penalty0 (9), 2019.
\newblock ISSN 2073-8994.
\newblock \doi{10.3390/sym11091066}.
\newblock URL \url{https://www.mdpi.com/2073-8994/11/9/1066}.

\bibitem[Medela \& Picon(2020)Medela and Picon]{constelationloss}
Alfonso. Medela and Artzai. Picon.
\newblock {Constellation loss: Improving the efficiency of deep metric learning
  loss functions for the optimal embedding of histopathological images}.
\newblock \emph{Journal of Pathology Informatics}, 11\penalty0 (1):\penalty0
  38, 2020.
\newblock \doi{10.4103/jpi.jpi_41_20}.
\newblock URL
  \url{https://www.jpathinformatics.org/article.asp?issn=2153-3539;year=2020;volume=11;issue=1;spage=38;epage=38;aulast=Medela;t=6}.

\bibitem[{Schroff} et~al.(2015){Schroff}, {Kalenichenko}, and
  {Philbin}]{TripletLossPaper}
F.~{Schroff}, D.~{Kalenichenko}, and J.~{Philbin}.
\newblock Facenet: A unified embedding for face recognition and clustering.
\newblock In \emph{2015 IEEE Conference on Computer Vision and Pattern
  Recognition (CVPR)}, pp.\  815--823, 2015.
\newblock \doi{10.1109/CVPR.2015.7298682}.

\bibitem[Simonyan \& Zisserman(2015)Simonyan and Zisserman]{VGG}
Karen Simonyan and Andrew Zisserman.
\newblock Very deep convolutional networks for large-scale image recognition.
\newblock In Yoshua Bengio and Yann LeCun (eds.), \emph{3rd International
  Conference on Learning Representations, {ICLR} 2015, San Diego, CA, USA, May
  7-9, 2015, Conference Track Proceedings}, 2015.
\newblock URL \url{http://arxiv.org/abs/1409.1556}.

\bibitem[Snell et~al.(2017)Snell, Swersky, and Zemel]{PrototipicalNetworks}
Jake Snell, Kevin Swersky, and Richard Zemel.
\newblock Prototypical networks for few-shot learning.
\newblock In I.~Guyon, U.~V. Luxburg, S.~Bengio, H.~Wallach, R.~Fergus,
  S.~Vishwanathan, and R.~Garnett (eds.), \emph{Advances in Neural Information
  Processing Systems}, volume~30, pp.\  4077--4087. Curran Associates, Inc.,
  2017.
\newblock URL
  \url{https://proceedings.neurips.cc/paper/2017/file/cb8da6767461f2812ae4290eac7cbc42-Paper.pdf}.

\bibitem[{Sung} et~al.(2018){Sung}, {Yang}, {Zhang}, {Xiang}, {Torr}, and
  {Hospedales}]{RelationNetwork}
F.~{Sung}, Y.~{Yang}, L.~{Zhang}, T.~{Xiang}, P.~H.~S. {Torr}, and T.~M.
  {Hospedales}.
\newblock Learning to compare: Relation network for few-shot learning.
\newblock In \emph{2018 IEEE/CVF Conference on Computer Vision and Pattern
  Recognition}, pp.\  1199--1208, 2018.
\newblock \doi{10.1109/CVPR.2018.00131}.

\bibitem[{Szegedy} et~al.(2015){Szegedy}, {Wei Liu}, {Yangqing Jia},
  {Sermanet}, {Reed}, {Anguelov}, {Erhan}, {Vanhoucke}, and
  {Rabinovich}]{Inception}
C.~{Szegedy}, {Wei Liu}, {Yangqing Jia}, P.~{Sermanet}, S.~{Reed},
  D.~{Anguelov}, D.~{Erhan}, V.~{Vanhoucke}, and A.~{Rabinovich}.
\newblock Going deeper with convolutions.
\newblock In \emph{2015 IEEE Conference on Computer Vision and Pattern
  Recognition (CVPR)}, pp.\  1--9, 2015.
\newblock \doi{10.1109/CVPR.2015.7298594}.

\bibitem[Vinyals et~al.(2016)Vinyals, Blundell, Lillicrap, kavukcuoglu, and
  Wierstra]{MatchingNetworkPaper}
Oriol Vinyals, Charles Blundell, Timothy Lillicrap, koray kavukcuoglu, and Daan
  Wierstra.
\newblock Matching networks for one shot learning.
\newblock In D.~Lee, M.~Sugiyama, U.~Luxburg, I.~Guyon, and R.~Garnett (eds.),
  \emph{Advances in Neural Information Processing Systems}, volume~29, pp.\
  3630--3638. Curran Associates, Inc., 2016.
\newblock URL
  \url{https://proceedings.neurips.cc/paper/2016/file/90e1357833654983612fb05e3ec9148c-Paper.pdf}.

\bibitem[{Wang} et~al.(2018){Wang}, {Wang}, {Zhou}, {Ji}, {Gong}, {Zhou}, {Li},
  and {Liu}]{NSL}
H.~{Wang}, Y.~{Wang}, Z.~{Zhou}, X.~{Ji}, D.~{Gong}, J.~{Zhou}, Z.~{Li}, and
  W.~{Liu}.
\newblock Cosface: Large margin cosine loss for deep face recognition.
\newblock In \emph{2018 IEEE/CVF Conference on Computer Vision and Pattern
  Recognition}, pp.\  5265--5274, 2018.
\newblock \doi{10.1109/CVPR.2018.00552}.

\end{thebibliography}
